\documentclass[conference]{IEEEtran}
\IEEEoverridecommandlockouts
\usepackage{cite}
\usepackage{amsmath,amssymb,amsfonts}
\usepackage{algorithmic}
\usepackage{graphicx}
\usepackage{subcaption}
\usepackage{longtable}
\usepackage{booktabs}
\usepackage{textcomp}
\usepackage{xcolor}
\usepackage{soul}

\usepackage{cleveref}
\crefformat{figure}{Fig.~#2#1#3}
\crefformat{section}{Sec.~#2#1#3}
\crefformat{appendix}{Appendix~#2#1#3}
\crefformat{table}{Table~#2#1#3}
\crefformat{equation}{Eq.~#2#1#3}

\usepackage[colorinlistoftodos,textsize=tiny,textwidth=10mm]{todonotes}

\begin{document}

\title{Reducing Inference Cost of Coding Agents with Timeout Cascades}

\author{
    \IEEEauthorblockN{Boxi Zhou}
    \IEEEauthorblockA{
        \textit{AMD Singapore} \\
    }
\and
    \IEEEauthorblockN{German Sviridov}
    \IEEEauthorblockA{
        \textit{AMD Singapore} \\
    }
\and
    \IEEEauthorblockN{Dmitrii Ustiugov}
    \IEEEauthorblockA{
        \textit{NTU Singapore} \\
    }
\and
    \IEEEauthorblockN{Haris Javaid}
    \IEEEauthorblockA{
        \textit{AMD Singapore} \\
    }
}

\maketitle

\begin{abstract}


AI coding agents have become one of the most popular applications of large language models. Yet, agentic AI comes at the cost of a tremendous amount of GPU compute resources, which limits its profitability and its viability at scale. In this paper, we investigate whether smaller, cheaper models can cut this cost without sacrificing accuracy. 

We evaluate 36 popular large and small open-weight models released since 2024 on a software engineering benchmark, and find that recent small models are closing the accuracy gap compared to their large counterparts. 
We investigate the structure of this gap and find that the gap is highly concentrated around hard tasks that small language models fail to solve while achieving remarkably good accuracy for easy and medium complexity tasks. Following this observation, we show that by utilizing a hybrid inference strategy that tries a small model first and escalates to a large one only when needed, we can achieve up to 33\% reduction in total GPU-hour costs and 16.4\% on all tasks from SWE-bench Verified, without sacrificing accuracy. Strikingly, this simple cascade nearly matches an oracle-style task router, highlighting a practical path for deploying cost-aware coding agents in production without complex learned task routers and with minimal engineering complexity.
\end{abstract}


\section{Introduction}

AI coding agents have quickly become one of the most practical use-cases of large language models (LLMs). Given a software issue, these agents can inspect a repository, edit code, run tests, and iterate until they produce a solution. This agentic workflow has pushed coding benchmark performance forward, but it also makes inference expensive: a single task may require many model calls, long execution traces, and substantial compute.

The common way to maximize a coding agent's accuracy is to run every task using the largest available model. This is simple and effective, but also wasteful. Many software tasks are routine enough for a smaller model, while the large model is only needed for harder cases. As coding agents move from demos to production systems, this distinction matters. The relevant question is no longer just whether an agent can solve a task, but how much inference it does to get there. Recent small open-weight models make this tradeoff worth revisiting.

\begin{figure}
\centering
\includegraphics[width=0.95\linewidth]{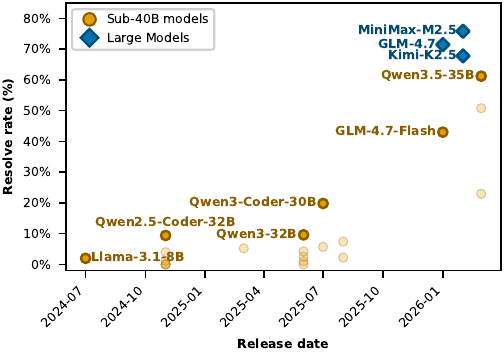}
\caption{
Agentic coding capability of selected models on SWE-bench Verified~\cite{jimenez2024swebench} using mini-swe-agent vs.\ release date. Full list of models and evaluation setup is described in \cref{sec:methodology}.
}
\label{fig:hero}
\end{figure}

\begin{figure}
    \centering
    \includegraphics[width=0.95\linewidth]{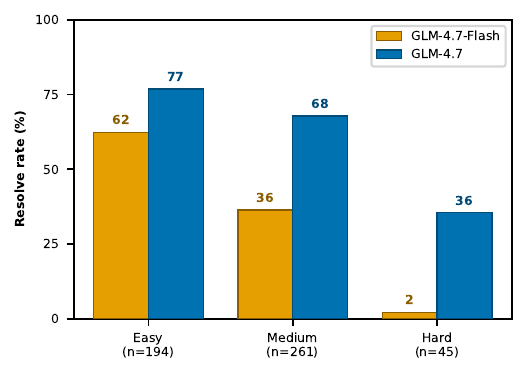}
    \caption{Resolve rate of GLM-4.7-Flash (30B) and GLM-4.7 (355B) on SWE-bench Verified~\cite{jimenez2024swebench}, broken down by task difficulty.}
    \label{fig:bar-difficulty-comparison}
\end{figure}

\cref{fig:hero} illustrates that sub-40B models (models with less than 40 billion total parameters) have improved from near-zero performance on SWE-bench Verified~\cite{jimenez2024swebench} to resolving a substantial fraction of real-world software engineering tasks under the same agent scaffold. However, the gap between small and large models has not disappeared. Instead, small models are competitive on easier tasks, weaker on medium tasks, and still struggle on hard tasks as shown in \cref{fig:bar-difficulty-comparison}. This suggests that small models should not replace large models outright, but they may be valuable as the first stage of a hybrid inference system.

In this paper we explore whether the observation above can be turned into a practical deployment strategy. We study 36 open-weight models released from mid-2024 through the first quarter of 2026 on SWE-bench Verified~\cite{jimenez2024swebench}, using a fixed mini-swe-agent scaffold and a consistent serving setup on AMD Instinct\textsuperscript{\texttrademark} MI300X GPUs. We then propose and evaluate two hybrid architectures: a timeout-based cascade, which tries a small model first and falls back to a large model only when needed, and a difficulty-aware task router, which uses task difficulty to decide which model to run first.

A timeout cascade architecture using GLM-4.7-Flash as the first-stage model and GLM-4.7 as the fallback preserves the large model's resolve rate while reducing measured inference cost. The reason is that execution time itself becomes a useful proxy: tasks the small model can solve tend to finish within the timeout, while tasks that exceed the budget naturally escalate to the larger model. This finding is important because the obvious alternative is a difficulty-aware task router that predicts, before execution, which tasks require the large model. Such routers can be effective, but they add training, calibration, and deployment complexity. In our evaluation, even an oracle-style task difficulty router provides only a small additional cost reduction over the timeout cascade. This suggests that, for current coding agent workloads, much of the value of routing can be captured by a far simpler execution-time rule.

The broader insight is that for many coding workloads, timeout is not merely a failure condition; it can also be an effective routing mechanism. Hence, building complex learned task routing infrastructure may be overkill. Note that the cascade architecture does not come for free. Some tasks are first attempted by the small model and then restarted on the large model, increasing wall-clock time. This makes the approach most attractive for asynchronous coding agent workflows, where tasks are delegated and reviewed later, rather than for interactive setups where users wait for each response. Our key contributions are as follows:

\begin{itemize}
    \item \textbf{Systematic evaluation and task-level analysis:} We evaluate 36 recent open-weight small and large models on SWE-bench Verified under a common mini-swe-agent scaffold, and analyze where small models succeed and fail relative to larger models. We show that recent sub-40B models have become credible first-stage options for coding agents, but that their remaining performance gap is concentrated in harder tasks.
    \item \textbf{Lightweight timeout cascade for cost-aware coding agents:} We propose and evaluate a timeout-based cascade architecture that runs a small model first and escalates to a large model only after failure or timeout. We compare this architecture against a difficulty-aware task router, showing that the cascade captures most of the router’s cost-saving benefit without requiring classifier training or additional routing infrastructure (16.4\% GPU-hour reduction from timeout cascade versus 17.9\% from task router).
    \item \textbf{Practical implications for deployment:} We show that combining small and large models within a single inference system can reduce GPU-hour cost by 16–18\% across all tasks, and up to 33\% on easy tasks, without sacrificing accuracy. These results suggest a practical path for deploying cost-aware coding agents in production, especially for asynchronous workflows where inference cost matters more than immediate response latency.
\end{itemize}

The rest of the paper is organized as follows. In \cref{sec:related-work}, we review the related work on coding agents, small models' capabilities, routing strategies, and agent economics. \cref{sec:methodology} describes our evaluation methodology, including the models evaluated, serving setup, benchmark and difficulty stratification, agent framework, cost measurement, and hybrid architecture simulation. In \cref{sec:results}, we present results and analysis for various research questions, while in \cref{sec:discussion} we describe the implications, limitations, and future directions based on the obtained results. Finally, \cref{sec:conclusion} presents the conclusions.

\section{Related Work}\label{sec:related-work}
In the following, we review related work on coding agents and agentic benchmarks, small models' capabilities, routing strategies for hybrid inference, and the economics of agent deployment.

\subsection{Coding Agents and SWE-bench}
Since its release, SWE-bench~\cite{jimenez2024swebench} has become the de facto benchmark for LLM-based coding agents. A succession of agent scaffolds (SWE-agent, OpenHands, and mini-swe-agent) has steadily raised resolve rates by equipping models with tools for repository navigation, test execution, and iterative repair~\cite{yang2024sweagent, wang2025openhandsopenplatformai}, with the state-of-the-art systems now exceeding 70\% on the verified subset. This literature, however, focuses almost exclusively on maximizing accuracy, treating inference cost as an afterthought. We instead keep the scaffold fixed and study how model selection and adaptive routing can cut inference cost without sacrificing the accuracy these scaffolds enable.

\subsection{Advances in Small Models' Capabilities}
While scaling laws long held that LLMs dominate across benchmarks, a growing body of work illustrates a rapid closing of this gap by small models, especially for software engineering tasks. A survey of 70 open-source small language models (SLMs) finds that they increasingly match larger counterparts on reasoning, mathematics, and coding~\cite{lu2025small}. Belcak et al. argue that many agentic invocations are repetitive, scoped, and non-conversational, and thus well suited to specialized SLMs; they report 2--9B models rivaling far larger ones on code generation, instruction following, and tool calling, and note that serving a 7B model can be 10--30$\times$ cheaper than a 70--175B LLM in latency, energy, and FLOPs~\cite{belcak2025small}. We extend this line of inquiry to full agentic software engineering workflows, asking not just how fast small models are improving, but whether their improvement \emph{across difficulty levels} enables practical efficiency strategies for coding agents.

\subsection{Cascades and Model Routing}

Routing queries across models of varying capability and cost, rather than relying on a single model, has been explored in several contexts. FrugalGPT introduced LLM cascades that invoke increasingly powerful models in sequence, stopping once a response meets a quality threshold~\cite{chen2023frugalgptuselargelanguage}. Subsequent work has explored learned routers that predict query difficulty from input features and dispatch each query to a model of appropriate capability and cost~\cite{ding2024hybridllm,ding2025bestroute,ong2025routellm}. In the coding domain, routing has been studied for generation and auto-repair, but primarily in single-turn setups.

Our work differs in three ways. We study routing for full agentic coding (multi-turn, tool-using, repository-scale tasks) rather than single-prompt generation; we provide what is, to our knowledge, the first head-to-head comparison of a timeout cascade and a difficulty-aware task router on the same benchmark; and, most importantly, our finding that the simple timeout cascade rivals the learned router challenges the common assumption that more sophisticated routing yields proportionate inference cost reductions.

\subsection{Agent Economics and Deployment Considerations}
As LLM agents move into production, cost becomes a first-order systems constraint. Coding agents are particularly expensive because they are not single-turn assistants anymore: each task can involve repeated repository inspection, patch generation, test execution, debugging, and repair. Empirical studies show that software engineering is already a high-volume AI workload, with usage concentrated in software development and writing tasks~\cite{handa2025economictasksperformedai}, and increasingly shifting from chat interfaces toward API-mediated and Claude Code workflows~\cite{anthropic2026aeiv5}. Industry analyses further emphasize that these multi-step loops make end-to-end cost per completed task a key deployment metric~\cite{blaxel2026agentcost}. This makes coding agents a natural target for cost-aware inference: at production scale, avoiding unnecessary large model calls can yield meaningful cost savings even when the underlying mechanism is simple (as demonstrated in this paper).

\section{Research Questions}

The prior work above points to a broad design space for combining small and large models. At one end are the \emph{cascades}, which start with a cheaper model and invoke a stronger model only when the first attempt fails or times out. At the other end are the \emph{routers}, which predict task difficulty or expected model success before execution and assign each task to an appropriate model. Other hybrid designs are also possible: parallel portfolios can race multiple models and accept the first valid solution; ensemble or voting systems can compare several candidate patches; speculative or assisted generation can use a small model to accelerate a large one; and budget-aware schedulers can allocate model calls under a global cost or latency constraint.

We focus on cascades and routers because they are the two most direct and deployable mechanisms for reducing a coding agent's cost without changing the underlying inference engine or model weights. They also represent complementary decision points. A router makes an \emph{up-front} decision from task features, while a cascade observes a small model's progress and uses its \emph{execution time} as the proxy for the task difficulty. This contrast lets us test whether explicit task difficulty prediction is necessary, or whether a simple timeout signal already captures most of the useful task routing decision. More complex, adaptive, and speculative systems remain promising, but they require additional acceptance criteria, parallel execution, reruns, or model-internal integration; we leave them as future work.

We structure the paper around five questions that connect model capability, task difficulty, system design, and deployment tradeoffs:

\textbf{RQ1: Can small models serve as first-stage coding agents?}
Before building hybrid inference systems, we first ask whether recent small and medium-sized open-weight models have become capable enough to run before a larger fallback model. In other words, are they credible first-stage agents for real software engineering tasks, or still cheaper but substantially weaker alternatives?

\textbf{RQ2: Where do small models still fall short?}
Aggregate resolve rates can obscure where the performance gap between small and large models comes from. We therefore ask whether small models fail uniformly across tasks, or whether their remaining weakness is concentrated in specific difficulty tiers. This distinction matters because a structured gap creates an opportunity for hybrid inference.

\textbf{RQ3: Can a small-model-first cascade reduce inference cost without losing accuracy?}
If small models can solve a meaningful share of routine tasks, a natural policy is to run them first and invoke the large model only when needed. We ask whether a timeout-based cascade can preserve the large model's resolve rate while reducing measured GPU-hour cost.

\textbf{RQ4: Do coding agents need difficulty-aware task routing?}
A task router could avoid wasted small-model attempts by predicting which tasks require the large model before execution. We ask how much additional cost reduction such a router provides over a simple timeout cascade, and whether that gain justifies the added training, calibration, and deployment complexity.

\textbf{RQ5: What latency cost does the cascade introduce?}
A small-model-first cascade can reduce inference cost, but it may increase wall-clock time because some tasks are first attempted by the small model and then restarted on the large model. We therefore ask how large this sequential-runtime overhead is, and which deployment settings can tolerate it.

\section{Methodology}\label{sec:methodology}

\begin{figure*}[!t]
    \centering
    \begin{minipage}{0.38\textwidth}
        \centering
        \includegraphics[width=0.76\linewidth]{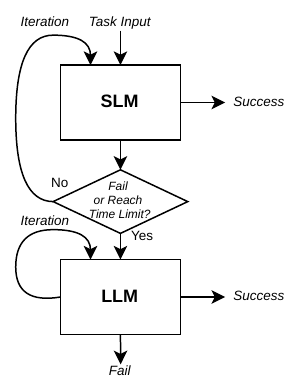}\\[-0.2em]
        \centerline{\footnotesize\textbf{(a) Timeout-based cascade}}
        \label{fig:cascade_architecture}
    \end{minipage}\hfill
    \begin{minipage}{0.56\textwidth}
        \centering
        \includegraphics[width=0.82\linewidth]{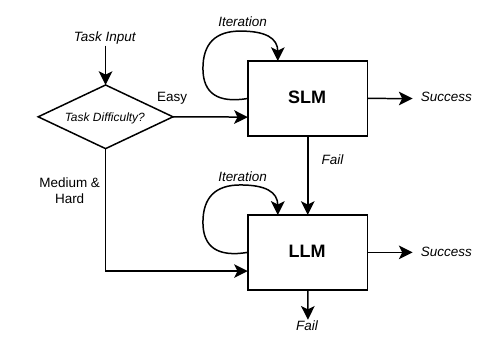}\\[-0.2em]
        \centerline{\footnotesize\textbf{(b) Difficulty-aware task router}}
        \label{fig:router_architecture}
    \end{minipage}
    \caption{Hybrid inference architectures evaluated by trace-driven simulation. The cascade starts every task on the smaller model and escalates after failure or timeout. The router makes an up-front decision from task difficulty, sending easy tasks to the smaller model and harder tasks to the larger model.}
    \label{fig:hybrid-architectures}
\end{figure*}

We split our evaluation strategy into two phases. First, we run each model on SWE-bench Verified~\cite{jimenez2024swebench} under the same mini-swe-agent scaffold and serving configuration, recording task outcomes, model trajectories, token counts, GPU usage and hours, and wall-clock runtime. Second, we use these single-model traces to simulate the two hybrid architectures shown in \cref{fig:hybrid-architectures}: a timeout-based cascade and a difficulty-aware task router. This separation keeps the agent evaluation fixed while varying only the policy that decides when to use a small or large model.

\subsection{Models Evaluated}

We evaluate 36 open-weight language models released from mid-2024 through the first quarter of 2026. The model suite spans both sub-40B and larger models, and covers code-specialized and general-purpose families including Qwen2, Qwen2.5, Qwen3, Qwen3.5, QwQ, Llama 3, GPT-OSS, GLM-4.7, Kimi-K2.5, and MiniMax-M2.5. All model checkpoints are sourced from their official Hugging Face repositories.

We use 40B parameters as a practical deployment cutoff, not a hard capability threshold. Models below this size are more likely to fit into cost-efficient serving setups (such as a single high-end datacenter or consumer GPU), making them plausible first-stage agents in a cascade. Models above this range may offer higher accuracy, but they are also more expensive to serve and more likely to require multi-GPU or distributed deployments. Therefore, in our setup, these larger models play the role of high-cost fallback or baseline setups rather than economical first-stage options. This distinction reflects our central systems question: can cheaper, easier-to-deploy models solve enough tasks before escalation to a larger model to reduce overall inference cost?

Note that some strong model families, including DeepSeek, are not included in our main benchmark suite. In pilot runs, these models required substantially longer wall-clock time per SWE-bench task, making a full evaluation impractical within our experimental budget. We therefore exclude them from the main results and discuss the resulting limitation in \cref{sec:discussion}.

\subsection{Serving and Hardware Setup}

We run all experiments using a vLLM-based serving stack on AMD Instinct\textsuperscript{\texttrademark} MI300X GPUs. Unless model compatibility requires otherwise, models are served with vLLM~0.16.0 on ROCm~7.0.0. The Qwen3.5 series is the only exception: it is served with vLLM~0.17.0 on ROCm~7.2.1.

For each model, we allocate the minimum number of GPUs required for a reliable deployment under our ROCm/vLLM setup. The allocation must both load the model checkpoint and reserve enough KV-cache capacity for long-context SWE-bench agent rollouts without out-of-memory failures. For multi-GPU deployments, we use vLLM tensor parallelism, setting the tensor-parallel degree equal to the number of allocated GPUs.


\subsection{Benchmark and Difficulty Stratification}\label{sec:difficulty-level}

We use SWE-bench Verified~\cite{jimenez2024swebench} in its entirety (500 tasks). Task difficulty labels are taken from the official SWE-bench dataset metadata, which provides human-annotated estimates of how long an experienced software engineer familiar with the codebase would need to resolve the issue. We group these labels into three categories:
\begin{itemize}
    \item \textbf{Easy}: estimated resolution time $<$ 15 minutes (194 tasks, 38.8\%).
    \item \textbf{Medium}: 15 minutes to 1 hour (261 tasks, 52.2\%).
    \item \textbf{Hard}: $>$ 1 hour (45 tasks, 9.0\%).
\end{itemize}

\subsection{Agent Framework}
\label{sec:agent-framework}

All models are evaluated with the same agent scaffold, \textbf{mini-swe-agent}, so that performance differences primarily reflect model behavior rather than differences in agent engineering. The mini-swe-agent is directly integrated with SWE-bench, supports batch evaluation on SWE-bench Verified, and uses a deliberately small scaffold with few hand-engineered heuristics. This makes it a suitable baseline for comparing models under a common software engineering environment.

Each SWE-bench task is initialized from its repository snapshot and issue description. The agent is allowed to interact with the repository through the fixed tool interface provided by mini-swe-agent, including shell execution, file inspection, code modification, and test execution. After the agent terminates, we extract the final patch generated by the model and evaluate it. A task is considered resolved only if the patched repository passes all unit tests specified by the SWE-bench evaluation for that instance.

We keep the agent prompt, tool interface, and execution limits fixed across all models. The main hyperparameters are:

\begin{itemize}
    \item \textbf{Maximum agent steps:} Fixed to 250. Each step corresponds to one model action and its associated tool execution or response cycle.
    \item \textbf{Maximum wall-clock time:} Fixed to 2 hours per task. Runs that exceed this limit are terminated and counted as unresolved.
    \item \textbf{Maximum generated tokens:} We do not impose a separate total-output-token budget beyond the model's context limit, the serving configuration, and the step/time limits above.
\end{itemize}

Each model-task pair is executed once due to the large scale of our evaluation. This follows common practice for SWE-bench-style agent evaluations, where full multi-seed evaluation across dozens of models is prohibitively costly.

\subsection{Cost Measurement}

We report per-task cost as the wall-clock GPU hours consumed on AMD Instinct\textsuperscript{\texttrademark} MI300X GPUs:
\begin{equation}
C_{\mathrm{gpu}} = t_{\mathrm{wall}} \times n_{\mathrm{gpu}}
\end{equation}
where $t_{\mathrm{wall}}$ is runtime in hours and $n_{\mathrm{gpu}}$ is the number of allocated GPUs. The resolve rate is measured as the fraction of submitted tasks whose final patch passes the SWE-bench verification tests. 

Our reported GPU-hour cost should be interpreted as an effective per-task cost under exclusive GPU allocation. In practice, we evaluate tasks in batches and allow up to 30 tasks to run in parallel to improve throughput. When computing the numbers reported in the figures, however, we charge each task by its own wall-clock runtime without amortizing the shared GPU usage across concurrently running tasks. As a result, the summed GPU-hours in the figures can be larger than the actual hardware GPU-hours consumed during evaluation, depending on the degree of parallelism. We use this convention to make per-task costs comparable across models and tasks, but the reported values should not be interpreted as the total physical GPU-hours used by the evaluation cluster.

Note that $t_{\mathrm{wall}}$ is directly proportional to the model complexity (e.g., its size or number of active parameters) and to the number of agent steps taken before termination. This means that a model that takes more steps to solve a task, even if it ultimately succeeds, will have a higher GPU-hour cost than a model that solves the same task in fewer steps.

\subsection{Trace-Driven Hybrid Policy Simulation}

For each model--task pair, the single-model evaluations record the agent trajectory, final outcome, token counts, GPU-hour cost, and wall-clock duration. We use these traces to simulate hybrid inference policies corresponding to the cascade and router architectures in \cref{fig:hybrid-architectures}, without rerunning the agent. This trace-driven setup isolates the policy question (i.e., how to combine small and large models) while holding the agent scaffold, task outcomes, and execution costs fixed.

For all hybrid policies, we use GLM-4.7-Flash as the small first-stage model and GLM-4.7 as the large fallback model. We choose this pair rather than the highest-resolve-rate sub-40B model (Qwen3.5-35B-A3B) in our evaluation because cascade savings depend on GPU-hour cost, and not the resolve rate alone. In our runs, the Qwen models required substantially more agent turns before termination, increasing their per-task wall-clock time and GPU-hour cost despite competitive resolve rates. GLM-4.7-Flash terminates in fewer turns, making it the more cost-efficient first stage, while pairing it with GLM-4.7 keeps the small and large models within the same family for a cleaner comparison.\footnote{The analysis in \cref{fig:hero} and \cref{fig:bar-difficulty-comparison} uses the full 500-task SWE-bench Verified set. The hybrid-policy simulations use the 466 tasks for which both GLM-4.7-Flash and GLM-4.7 produced complete recorded trajectories: 190 easy, 238 medium, and 38 hard. The remaining tasks lacked a usable trajectory for at least one of the two models.}

When a task falls back, the large model is treated as a cold restart: it begins from the original task state rather than inheriting the small model's partial edits, tool outputs, or intermediate reasoning. The total policy cost is computed as the sum of GPU-hours across all invoked model attempts. This assumption is conservative relative to systems that could reuse first-stage state, which we leave as a future optimization.

\subsection{Timeout-Based Cascade Architecture}
The timeout-based cascade, illustrated in \cref{fig:cascade_architecture}, routes every task first to GLM-4.7-Flash under a configurable time limit. If the small model produces a valid patch within this budget and the patch passes all task tests, the task is considered solved and only the small-model attempt is counted. Otherwise, if the run times out, produces an empty patch, introduces a syntax error, or fails the tests, the task is escalated to GLM-4.7, which restarts from the original task state. The final GPU-hour cost is the sum of the invoked model attempts. This policy tests whether a simple small-model-first execution rule can preserve the large model's resolve rate while avoiding unnecessary large-model calls on easier tasks.

We sweep the small-model timeout from 0 to 120 minutes, matching the overall two-hour wall-clock limit, and report how resolve rate and GPU-hour cost change with the timeout budget. This sweep identifies the cost--accuracy tradeoff and highlights the Pareto-optimal region.

\subsection{Difficulty-Aware Task Router Architecture}

The difficulty-aware task router, illustrated in \cref{fig:router_architecture}, dispatches tasks based on predicted task difficulty. Rather than training a classifier, we simulate router behavior using the ground-truth difficulty labels from \cref{sec:difficulty-level}. This lets us sweep routing accuracy from 0\% (entirely inaccurate dispatch) to 100\% (perfect prediction) and estimate an empirical upper bound on the efficiency gains achievable by difficulty-aware routing.

At 100\% accuracy, the router sends easy and medium tasks to GLM-4.7-Flash and hard tasks directly to GLM-4.7. Tasks routed to GLM-4.7-Flash that fail are escalated to GLM-4.7 under the same cold-restart assumption used by the timeout-based cascade architecture. We evaluate the router on the same task set as the cascade and compute resolve rate and GPU-hour cost at each routing-accuracy level. We discuss the challenges of training and deploying a real task difficulty classifier in \cref{sec:discussion}.

\section{Results and Analysis}\label{sec:results}

\begin{figure*}[t]
    \centering
    \begin{subfigure}{0.48\textwidth}
        \centering
        \includegraphics[width=\linewidth]{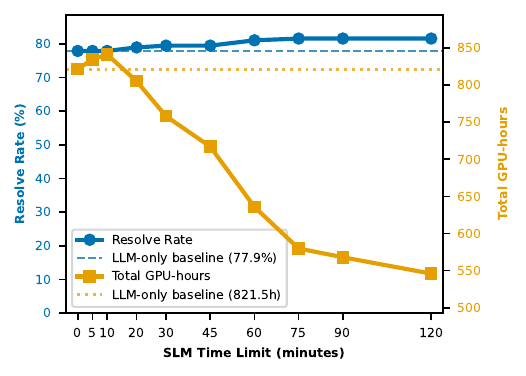}
        \caption{Easy tasks only}
        \label{fig:cascade-dual-easy}
    \end{subfigure}\hfill
    \begin{subfigure}{0.48\textwidth}
        \centering
        \includegraphics[width=\linewidth]{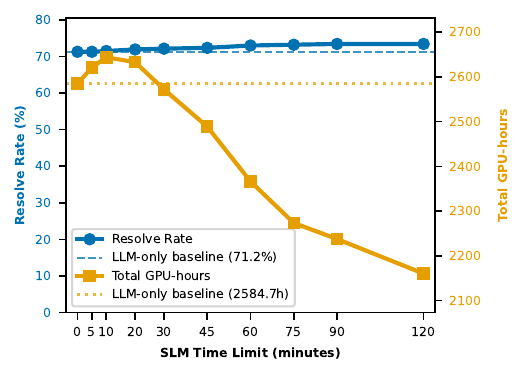}
        \caption{All tasks}
        \label{fig:cascade-dual-all}
    \end{subfigure}
    \caption{Timeout-based cascade's resolve rate (left axis) and total GPU-hour cost (right axis) vs.\ the GLM-4.7-Flash timeout, with the GLM-4.7-only baseline shown as dashed lines.}
    \label{fig:cascade-dual}
\end{figure*}

We now answer the research questions posed earlier. The first two questions establish whether small models are capable enough, and on which tasks. The remaining questions evaluate whether that capability can be converted into lower-cost coding-agent systems without sacrificing overall resolve rate.

\subsection{RQ1: Small Models Have Become Credible First-Stage Agents}\label{sec:growth}

\cref{fig:hero} answers RQ1 by showing how the SWE-bench Verified resolve rate of evaluated sub-40B models has changed over release time, with GLM-4.7, Kimi-K2.5, and MiniMax-M2.5 included as large-model references. The trend is clear: models below 40B parameters have moved from near-zero resolve rates to 61.2\% for the strongest recent model. The leading sub-40B model, Qwen3.5-35B-A3B, is no longer merely a cheap but weak alternative; under the same agent scaffold, it solves a substantial fraction of real software engineering tasks. This does not mean that small models can replace the largest models outright. Rather, it establishes the premise for hybrid inference: recent small models are now capable enough to serve as useful first-stage agents.

\subsection{RQ2: The Small--Large Gap Is Concentrated in Hard Tasks}

Aggregate resolve rates do not reveal where small models are useful, or where they still fail. \cref{fig:bar-difficulty-comparison} answers RQ2 by comparing GLM-4.7-Flash and GLM-4.7 across task difficulty levels. On easy tasks, the small model resolves 62\% of issues compared with 77\% for the large model, reaching 81\% of the large model's resolve rate. On medium tasks, the gap is wider: 36\% versus 68\%, or 53\% relative performance. On hard tasks, the gap becomes much larger, with the small model resolving only 2\% of the tasks compared with 36\% for the large model in our run.

This difficulty gradient is what makes hybrid inference attractive. Easy and medium tasks account for 91\% of SWE-bench Verified, or 455 out of 500 tasks. If many of these tasks can be attempted cheaply by a small model, while failures and harder cases fall back to a larger model, a hybrid system can potentially preserve much of the large model's final accuracy at lower GPU-hour cost.

\subsection{RQ3: A Small-Model-First Cascade Reduces GPU-Hour Cost Without Sacrificing Accuracy}
\label{sec:result-cascade}

\cref{fig:cascade-dual-easy} and \cref{fig:cascade-dual-all} answer RQ3 by sweeping the GLM-4.7-Flash timeout from 0 to 120 minutes and reporting the resulting resolve rate and GPU-hour cost. The sweep reveals a clear tradeoff:

\textbf{Ultra-short timeouts}: With very short timeouts, the small model is often stopped before it can solve tasks it would otherwise complete, causing unnecessary fallback to the large model.

\textbf{Sufficient timeouts}: As the timeout increases, more easy and medium tasks are resolved by the cheaper first-stage model, reducing total GPU-hour cost.

The largest cost reduction occurs at the 120-minute timeout, the upper end of our sweep. On easy tasks, GPU-hour cost drops from 821 to 546 GPU-hours, a 33.5\% reduction. Across all tasks, it drops from 2585 to 2160 GPU-hours, a 16.4\% reduction. Crucially, these savings do not come from lowering the success criterion or accepting lower-quality outputs: the cascade preserves the large-model-only resolve rate.

The cascade also slightly exceeds the large-model-only resolve rate. This occurs because a small number of tasks that GLM-4.7 fails are solved by GLM-4.7-Flash. Manual inspection suggests a possible complementarity pattern: in some of these cases, the larger model appears to over-engineer the fix, while the smaller model applies a more direct patch. We interpret this complementarity cautiously. Because each model--task pair is evaluated once, some of the effect may reflect run-to-run variance rather than a stable difference between the models. We therefore treat the main result as cost reduction without accuracy loss, and leave multi-run validation of model complementarity to future work.

\subsection{RQ4: A Simple Timeout-Based Cascade Nearly Matches an Oracle-Style Task Router}\label{sec:router}

\begin{figure}
    \centering
    \includegraphics[width=0.95\linewidth]{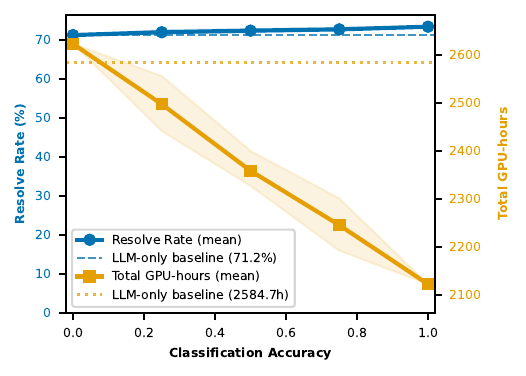}
    \caption{
        Router resolve rate and GPU-hour cost as routing accuracy is swept on all tasks.
        GLM-4.7-Flash is used as the small model and GLM-4.7 as the large model. The simulated router ranges from 0\% routing accuracy to 100\% oracle-style dispatch. At 100\%, easy and medium tasks are routed to the small model, while hard tasks are routed directly to the large model. Tasks routed to the small model that fail are escalated to the large model under the cold-restart assumption. Shades indicate the maximum observed deviation from the mean across routing seeds.
        }
    \label{fig:router-dual-all}
\end{figure}

\cref{fig:router-dual-all} answers the first part of RQ4 by sweeping the accuracy of the simulated difficulty-aware router from 0\% to 100\% and measuring resolve rate and GPU-hour cost. At 0\% accuracy, routing decisions are entirely incorrect under our sweep, and total cost rises to 2623 GPU-hours, slightly above the large-model-only baseline of 2585 GPU-hours. As routing accuracy improves, GPU-hour cost declines because more easy and medium tasks are assigned to the small model.

At 100\% accuracy, the router perfectly sends easy and medium tasks to GLM-4.7-Flash and hard tasks directly to GLM-4.7. This reduces cost from 2585 to 2122 GPU-hours, a 17.9\% reduction. As with the cascade, the overall resolve rate slightly exceeds the 71.2\% large-model-only baseline, due to the same complementarity effect discussed in \cref{sec:result-cascade}.

\begin{figure}
    \centering
    \includegraphics[width=0.95\linewidth]{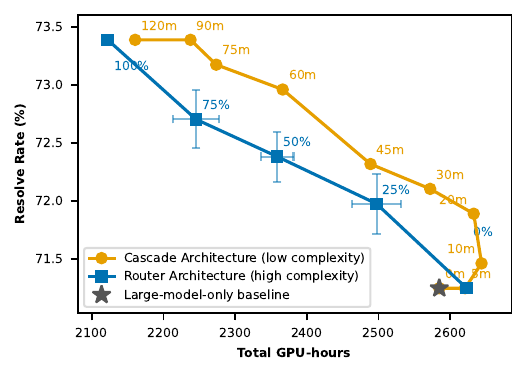}
    \caption{
        Cost--Accuracy comparison of the timeout cascade and oracle-style task router using GLM-4.7-Flash as the small model and GLM-4.7 as the large model. Upper left is better. Cascade points sweep the small-model timeout; router points sweep routing accuracy. Error bars indicate the maximum observed deviation from the mean across routing seeds.
        }
    \label{fig:comparison}
\end{figure}

\cref{fig:comparison} and \cref{tbl:comparison} answer the second part of RQ4 by comparing the router against the timeout cascade. The key result is that the cascade captures most of the router's benefit with far less complexity. The best cascade point reduces GPU-hour cost by 16.4\%, while the oracle-style task router reaches 17.9\%, an additional gain of only 1.5\%. Given the difficulty of training a reliable task-difficulty classifier for unseen coding problems, this small gap suggests that the timeout cascade is the stronger default: it achieves nearly the same cost reduction without classifier training, calibration, or routing infrastructure.

\begin{table}
    \centering
    \footnotesize
    \setlength{\tabcolsep}{4pt}
    \caption{Comparison of Cascade and Router architectures.}
    \begin{tabular}{lrrl}
        \toprule
        \textbf{Architecture} & \textbf{Resolve} & \textbf{GPU-Hour} & \textbf{Complexity} \\
         & \textbf{Rate} & \textbf{Reduction} &  \\
        \midrule
        Large-model-only (baseline) & 71.2\% & N/A & N/A \\
        Timeout cascade (best timeout) & 73.4\% & 16.4\% & Low \\
        Task router (oracle-style) & 73.4\% & 17.9\% & High \\
        \bottomrule
    \end{tabular}
    \label{tbl:comparison}
\end{table}

\subsection{RQ5: Cascade Savings Trade GPU-Hour Cost for Wall-Clock Time}\label{sec:wall-time-result}

\begin{figure}[t]
    \centering
    \includegraphics[width=0.95\linewidth]{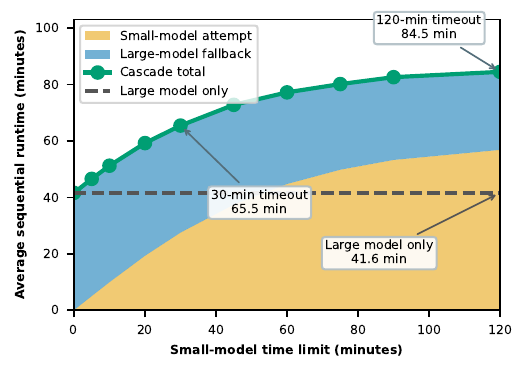}
    \caption{Average sequential runtime per SWE-bench Verified task under the GLM-4.7-Flash $\rightarrow$ GLM-4.7 cascade. The orange area shows time spent by the small model before success or escalation; the blue area shows time spent by the large model on escalated tasks; the green line shows total runtime.}
    \label{fig:wall-time}
\end{figure}

\cref{fig:wall-time} answers RQ5 by quantifying the wall-clock overhead of the small-model-first cascade. We measure total task runtime, including both model inference and agent scaffold execution. In our runs, tool execution accounts for less than 5\% of total execution time, so the observed overhead is dominated by model inference. For any positive small-model timeout, the cascade is slower than the large-model-only baseline because some tasks are first attempted by GLM-4.7-Flash and then restarted from scratch on GLM-4.7. The large-model-only baseline averages 41.6 minutes per task; the cascade averages 65.5 minutes with a 30-minute small-model timeout and 84.5 minutes with a 120-minute timeout.

This result clarifies the deployment tradeoff. The cascade reduces GPU-hour cost and preserves resolve rate, but it does so by spending additional sequential time on tasks that eventually fall back. This makes the approach most attractive for asynchronous coding-agent workflows, such as Google Jules and Open SWE~\cite{google2025jules, langchain2025openswe}, where tasks are delegated, executed in the background, and reviewed later. It is less suitable for interactive settings, such as conversational debugging or pair programming, where users wait for each response.

\section{Discussion and Future Work}\label{sec:discussion}
We now discuss why the timeout cascade works, where its benefits may or may not transfer, and what remains open for future work.

\subsection{Interpreting the Rapid Rise of Small Models}

The rapid improvement of small models on SWE-bench (\cref{sec:growth}) raises a natural question: what is driving this shift? Our study is not designed to isolate causal factors, but several trends likely contribute. First, post-training methods, including reinforcement learning with execution feedback, have improved substantially, helping smaller models internalize and perform complex multi-step reasoning that was previously associated mainly with larger models. Second, training data curation and synthetic code data generation have matured, giving smaller models more high-quality training data. Third, this trend appears across multiple open-weight model families, suggesting a broader advancement in training practice rather than a single model-specific breakthrough.

The practical implication is more important than the exact cause: the cost-efficiency frontier for coding agents is moving toward smaller models. Small models are still not replacements for the largest models on the hardest tasks, but they have become capable enough to absorb a meaningful share of routine software engineering workloads.

\subsection{Why Does a Simple Cascade Nearly Match an Oracle-Style Task Router?}

The most striking result of our study is the near-parity between the timeout cascade and the oracle-style task router (\cref{sec:router}). At first glance, this is surprising: a router with access to ground-truth task difficulty should have a clear advantage over a simple timeout rule. In practice, the answer lies in two properties of the workload.

First, SWE-bench Verified is heavily skewed toward easy and medium tasks. These tasks account for 91\% of the benchmark, creating a large opportunity for small-model-first execution. A perfect task router exploits this structure by sending easy and medium tasks to GLM-4.7-Flash and hard tasks directly to GLM-4.7, reducing GPU-hour cost by 17.9\% relative to the GLM-4.7-only baseline. However, the timeout cascade already captures most of this gain: it gives the small model a chance to solve the easier cases, while naturally escalating tasks that exceed the time budget.

Second, timeout acts as an implicit difficulty signal. Tasks that GLM-4.7-Flash can solve tend to finish within a bounded window; tasks that it cannot solve often consume the budget and fall back. This signal is not perfect, but it is available at near-zero cost: it requires no training data, no classifier calibration, and no additional routing infrastructure. This explains why the cascade performs similarly to a router with high routing accuracy. Much of the useful routing information is already exposed during execution.

This does not mean task routers are unnecessary in general. Explicit routing may provide larger gains when the task distribution is harder, when failed small-model attempts are very expensive, or when richer workload-specific signals are available before execution. But for SWE-bench-like coding workloads, where many tasks are routine and well specified, the timeout cascade is a strong default and its simplicity is hard to beat.

\subsection{Model Complementarity: Real Effect or Stochastic Artifact?}

A small but noteworthy result is that both hybrid policies slightly exceed the large-model-only baseline in overall resolve rate (\cref{sec:result-cascade}). Manual inspection suggests one possible explanation: in some cases, GLM-4.7 appears to over-engineer the fix, introducing unnecessary changes that fail tests, while GLM-4.7-Flash applies a simpler patch that passes. This observation is consistent with recent work on overthinking in reasoning models, which shows that additional test-time reasoning is not always beneficial and can sometimes move models away from correct solutions~\cite{zhou2026thinkinghurtsoverthinkingllm,wan2026mitigatingoverthinkinglargereasoning}.

We treat this complementarity cautiously. Each model--task pair is evaluated once, so part of the observed effect may reflect run-to-run variance rather than a stable difference between the models. The trajectories we inspected are suggestive, not conclusive. Establishing whether this complementarity is reliable would require multi-run evaluation across models and tasks, which we leave to future work. Our conservative conclusion is therefore narrower: adding a small model before the large model does not degrade overall accuracy in our setting; any complementary gains are an encouraging but tentative and not-fully-verified bonus.

\subsection{When Does a Cascade Help, and When Might It Fail?}

The timeout cascade is effective under our experimental conditions, but its success depends on two key preconditions.

\textbf{Easy and medium tasks must form a large share of the workload.}
In SWE-bench Verified, 91\% of tasks fall into these categories. The cascade saves GPU-hour cost because the small model can resolve many of these tasks without escalation. If a deployment workload were dominated by hard tasks, most first-stage attempts would fail, and the cascade would approach a large-model-only system with extra overhead from unsuccessful small-model runs. A related research question is whether \emph{hard and medium tasks can be decomposed into easier subtasks} that the small model can solve, effectively shifting the difficulty distribution in favor of the small-model-first cascade.

\textbf{The small model must be meaningfully more cost-efficient than the large model.}
Cost-efficiency requires both sufficient capability and a lower serving footprint. In our study, GLM-4.7-Flash reaches 81\% of GLM-4.7's resolve rate on easy tasks and 53\% on medium tasks, while using fewer GPU resources. This combination makes it useful as a first-stage model. If the small model were nearly as expensive to serve as the large model, or too weak to solve routine tasks reliably, the cascade would lose its economic advantage.

In practice, both preconditions are likely met in many real-world coding agent deployments, where routine bug fixes, test repairs, and localized feature requests dominate and cost-efficient small models are increasingly available. However, for domains with uniformly high difficulty or where small models lack the requisite capability, direct large-model invocation or more sophisticated routing may be preferable.

These preconditions also highlight a limitation of our study: we evaluate one primary model pair. Other pairs, or multi-stage cascades with more than two models, may yield different cost--accuracy tradeoffs. A stronger first-stage model could reduce escalation frequency, while a cheaper but weaker model could improve cost only if its failures are inexpensive. Finding the \textbf{optimal model combination} is therefore an important direction for future work. Likewise, speculative decoding is another complementary direction. Unlike cascades, which decide whether to invoke a large model at all, speculative decoding reduces generation cost by coupling a target model with a draft model~\cite{leviathan2023fast}. This can improve serving efficiency, but it addresses a different layer of the system. Combining speculative decoding with small-model-first cascades may provide additional savings, but it also introduces tighter coupling between models and reduces deployment flexibility.

\subsection{Practical Implications}

Our findings suggest several practical lessons for deploying coding agents.

\textbf{Use small-model-first execution when GPU-hour cost is the binding constraint.}
For SWE-bench-like workloads, trying a cheaper model first with a generous timeout is a low-complexity way to reduce GPU-hour cost while preserving the safety of a large-model fallback. This recommendation is strongest for asynchronous workflows, where tasks are delegated, executed in the background, and reviewed later.

\textbf{Tune the timeout as a cost--latency knob.}
Longer timeouts give the small model more opportunity to solve tasks and avoid large-model calls, but they also increase sequential wall-clock time for tasks that eventually fall back. Operators should therefore choose the timeout based on whether the deployment prioritizes GPU-hour cost, turnaround time, or a balance of both.

\textbf{Characterize the task distribution before training routers.}
The value of task routing depends heavily on the workload. If most tasks are routine, a timeout cascade may capture much of the available cost reduction without learned routing. If tasks are mostly hard or highly heterogeneous, a difficulty-aware task router may be worth the additional complexity. Measuring the deployment task mix should therefore come before investing in classifier training or routing infrastructure.

\textbf{Treat the current results as a strong but bounded case study.}
Our conclusions are constrained by single-run evaluations, one primary model pair, and one benchmark. The central design lesson is not that cascades always dominate routers, but that a simple timeout cascade is a strong first baseline. More complex routing systems should demonstrate gains beyond this baseline that justify their added training, calibration, and operational cost.

\section{Conclusion}\label{sec:conclusion}
In this paper, we study whether small language models can serve as practical first-stage agents in agentic coding systems. We evaluate 36 open-weight language models on SWE-bench Verified and analyze their tradeoffs in resolve rate and GPU-hour cost. We then simulate two hybrid inference policies that combine small and large models: a timeout-based cascade and a difficulty-aware task router. Both policies preserve the large-model resolve rate while reducing GPU-hour cost by 16--18\% across all tasks and up to 33\% on easy tasks, but the cascade achieves these savings with minimal engineering overhead. Our results suggest that a simple timeout cascade is a strong default for cost-aware coding agents, especially when workload-specific task routing signals are unavailable. More broadly, small models have shifted the cost-efficiency frontier for agentic coding: when used as first-stage agents, they can reduce inference cost without sacrificing final accuracy.

\bibliographystyle{IEEEtran}
\bibliography{biblio}

\end{document}